\documentclass[11pt, conference]{IEEEtran}
\IEEEoverridecommandlockouts
\usepackage{cite}
\usepackage{amsmath,amssymb,amsfonts}
\usepackage{algorithmic}
\usepackage{graphicx}
\usepackage{textcomp}
\usepackage{xcolor}
\usepackage{multirow}
\usepackage{hyperref}
\usepackage{booktabs}
\def\BibTeX{{\rm B\kern-.05em{\sc i\kern-.025em b}\kern-.08em
    T\kern-.1667em\lower.7ex\hbox{E}\kern-.125emX}}
\begin{document}

\title{Visible to Thermal image Translation for improving visual task in low light conditions}

\author{\IEEEauthorblockN{Md Azim Khan}
\IEEEauthorblockA{\textit{Department of Information Systems} \\
\textit{University of Maryland, Baltimore County (UMBC)}\\
azimkhan22@umbc.edu}
}

\maketitle

\begin{abstract}
Several visual tasks, such as pedestrian detection and image-to-image translation, are challenging to accomplish in low light using RGB images. Heat variation of objects in thermal images can be used to overcome this. In this work, an end-to-end framework, which consists of a generative network and a detector network, is proposed to translate RGB image into Thermal ones and compare generated thermal images with real data.
We have collected images from two different locations using the Parrot Anafi Thermal drone. After that, we created a two-stream network, preprocessed, augmented, the image data, and trained the generator and discriminator models from scratch. The findings demonstrate that it is feasible to translate RGB training data to thermal data using GAN. As a result, thermal data can now be produced more quickly and affordably, which is useful for security and surveillance applications.
\end{abstract}


\section{Introduction}
\label{intro}
Application of visual tasks are challenging in low light scenarios due to loss of information. As well as, it is difficult to take such pictures without specialized equipment \cite{liu2022survey}. Thermal camera which give heat data of objects can be used. The main issue with thermal cameras is their high cost as compared to visible cameras, which are widely available and even included in smartphones \cite{nielsen2014taking}. But Visible cameras and everyday use cameras lack the optical and sensor capabilities necessary to capture high-resolution images at night.The images taken by RGB are often fuzzy or have noticeable grains or noises, making them practically unusable for security-related applications \cite{haglund2016infrared}.

The objective is to generate thermal images of the visible images using cycle-consistent generative adversarial network \cite{goodfellow2020generative} (GAN).The domain translation is performed using RGB image. A cycle-consistent gan includes two neural network models referred to as generators \cite{kalal2010forward}. The goal of the first generator is to learn a mapping from the RGB domain to the authentic and the other one to learn a mapping for the other way around. For tackling this problem, we are going to use RGB, Thermal image dataset that are taken with the help of a drone and it has been ensured that the images are consistent with space and time. 

\section{Related Work}
\label{rw}

There are much fewer works investigating RGB–thermal image translation than other image translation due to the larger modality gap between the thermal and visible images \cite{wang2013heterogeneous}. Conventional translation methods perform image-to-image translation using feature-level adaptation methods\cite{long2015learning}. Translation of this kind typically involves minimising some measure of distance between the source and the target feature distributions, e.g. as in maximum mean discrepancy \cite{sutherland2016generative}, adversarial discriminator accuracy \cite{sun2016deep}or correlation distance \cite{tzeng2017adversarial}. These methods have successfully been used for semi supervised or unsupervised domain adaptation tasks\cite{sutherland2016generative,tzeng2017adversarial}. The weakness of these methods is that they do not enforce semantic consistency. Further, alignment at a higher level of a deep representation may not create alignment at a lower level of representation, which may be crucial for the final mapping.

More recently, generative adversarial networks (GANs) based methods are also worth noticing. here J.Goodfellow et al.\cite{goodfellow2020generative} adopt an adversarial loss to learn the mapping such that the translated images cannot be distinguished from images in the target domain. Generative pixel-level adaptation models instead perform distribution alignment in the pixel space instead of the feature space. This includes the methods pix2pix \cite{isola2017image},  bicycleGAN \cite{zhu2017toward} and MUNIT \cite{huang2018multimodal}.  Both the pix2pix\cite{isola2017image} and the bicycleGAN\cite{zhu2017toward} approach require paired training examples. Both uses a conditional generative adversarial network to learn a mapping from input to output images in supervised way which is favourable in many image-to-image translation tasks. Especially, if the outputs are highly structured and graphical. 

However, CycleGAN \cite{zhu2017unpaired}, an unpaired image translation method,introduced cycle consistency into the model, which can learn a bidirectional translation between two image distributions.Author exploit the property that translation should be “cycle consistent”, in the sense that if we translate a sentence from English to French, and then translate it back from French to English, we should arrive back at the original sentence. In this work, CycleGAN based RGB to Thermal image translation will be implemented using a dataset collected by drone.

\section{Methodology}
\label{met}
\subsection{Model Objective function}
Our objective contains two terms just as in \cite{zhu2017unpaired}.One is adversarial losses which is used for matching the distribution of generated images to the data distribution in the target domain.

\begin{figure}[!http]
\centering
\includegraphics[width=\linewidth]{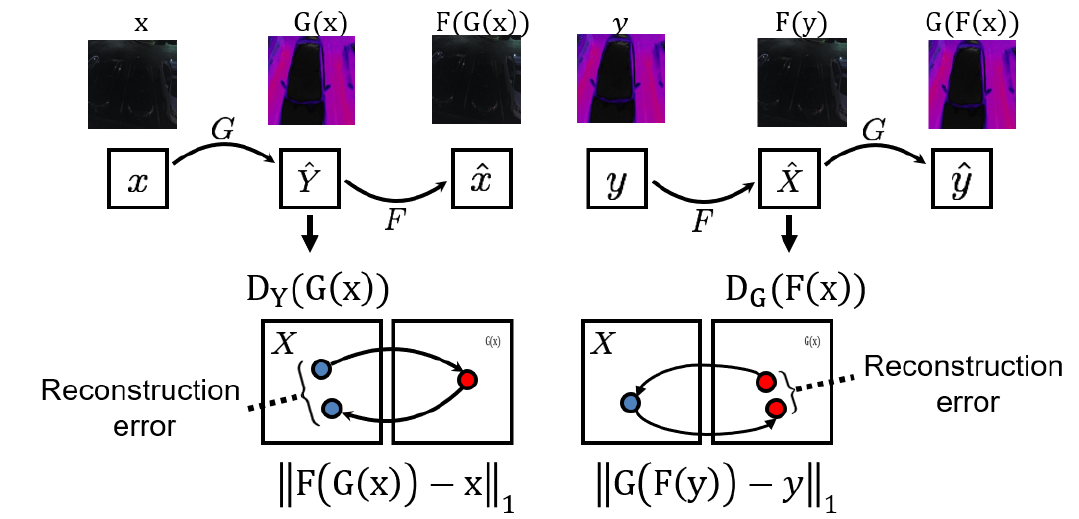}
\caption{Model objective function}
\label{obj}
\end{figure}

Let X be the domain of RGB images and Y the domain of Thermal shown in Figure \ref{obj}, where each sample $x\in X$ is a RGB image dataset. For the mapping function $G:\ X\rightarrow\ Y$ and its discriminator $D_Y$ , the objective function is

$$
\begin{gathered}
\mathcal{L}_{\mathrm{GAN}}\left(G, D_Y, X, Y\right)=E_{y \sim p_Y(y)}\left[\log D_Y(y)\right] \\
+E_{x \sim p_X(x)}\left[\log \left(1-D_Y(G(x))\right)\right]
\end{gathered}
$$

where G attempts to generate images G(x) that are similar to images from domain Y and 
$D_Y$ tries to distinguish between real samples y and generated samples G(x). Thus G tries to minimize the objective while $D_Y$ tries to maximize it resulting in
$\min _G \max _{D_Y} \mathcal{L}_{G A N}\left(G, D_Y, X, Y\right)$. 
Similarly, the adversarial function for the mapping function $F:\ Y\rightarrow\ X$  its discriminator $D_X$ is $\min _F \max _{D_X} \mathcal{L}_{G A N}\left(F, D_X, Y, X\right)$.

Another loss function is cycle consistency loss  as adversarial losses alone cannot guarantee that the learned function can map an individual input x to a desired output y \cite{zhu2017unpaired}.This expressed as

$$
\begin{gathered}
\mathcal{L}_{\text {cyc }}(G, F)=\mathbb{E}_{x \sim p_X(x)}\left[\|F(G(x))-x\|_1\right]+ \\
\mathbb{E}_{y \sim p_Y(y)}\left[\|G(F(y))-y\|_1\right]
\end{gathered}
$$

The full objective consists of adversarial loss and cycle consistency loss which can be written as:

$$
\begin{gathered}
\mathcal{L}\left(G,F,D_X,D_Y\right) \&=\mathcal{L}_{GAN}\left(G,D_Y,X,Y\right)\ +\\ \mathcal{L}_{GAN}\left(F,D_X,Y,X\right)+\lambda\mathcal{L}_{cyc}(G,F)
\end{gathered}
$$

Model cycle consistency and adversarial losses illustrated in Figure \ref{obj} where generator G mappings RGB to Thermal and Generator F mappings Thermal to RGB with their discriminator.

\subsection{Generator framework}

\begin{figure}[!http]
\centering
\includegraphics[width=\linewidth]{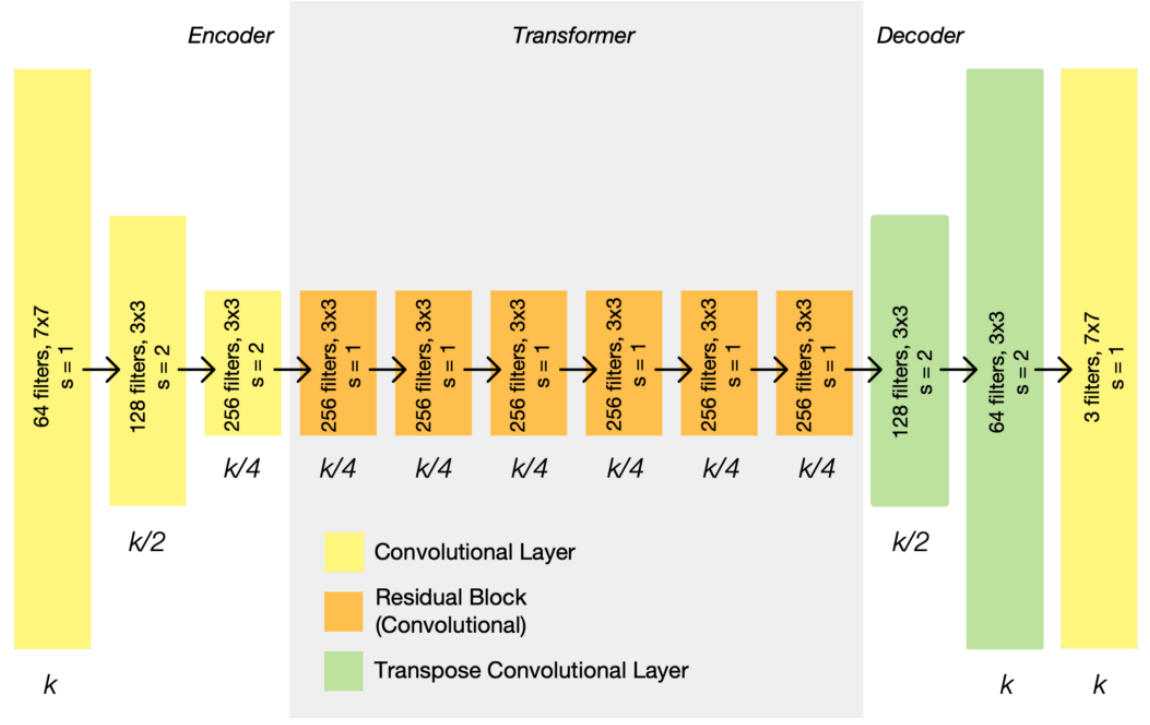}
\caption{generator architecture}
\label{dataset}
\end{figure}

Generator architecture for our  networks is taken from Johnson et al.\cite{johnson2016perceptual}who have shown impressive results for neural style transfer and super-resolution. Each CycleGAN generator consists of an encoder, a transformer, and a decoder shown in Figure \ref{dataset}. The input image is fed directly into the encoder, which shrinks the representation size while increasing the number of channels. The encoder is composed of three convolution layers. The resulting activation is then passed to the transformer, a series of six residual blocks \cite{he2016deep}. It is then expanded again by the decoder, which uses two transpose convolutions to enlarge the representation size, and one output layer to produce the final image in RGB.
The representation size that each layer outputs is listed in terms of the input image size, k Figure \ref{dataset}. On each layer is listed the number of filters, the size of those filters, and the stride with $\frac{1}{2}$. Each layer is followed by an instance normalization\cite{xu2018effectiveness} and ReLU activation\cite{agarap2018deep} ,2 fractionally-stride convolutions with stride $\frac{1}{2}$ , and one convolution that maps features to RGB. We use  9 blocks for 256×256 and higher-resolution training images.

\subsection{Discriminator framework}

\begin{figure}[!http]
\centering
\includegraphics[width=\linewidth]{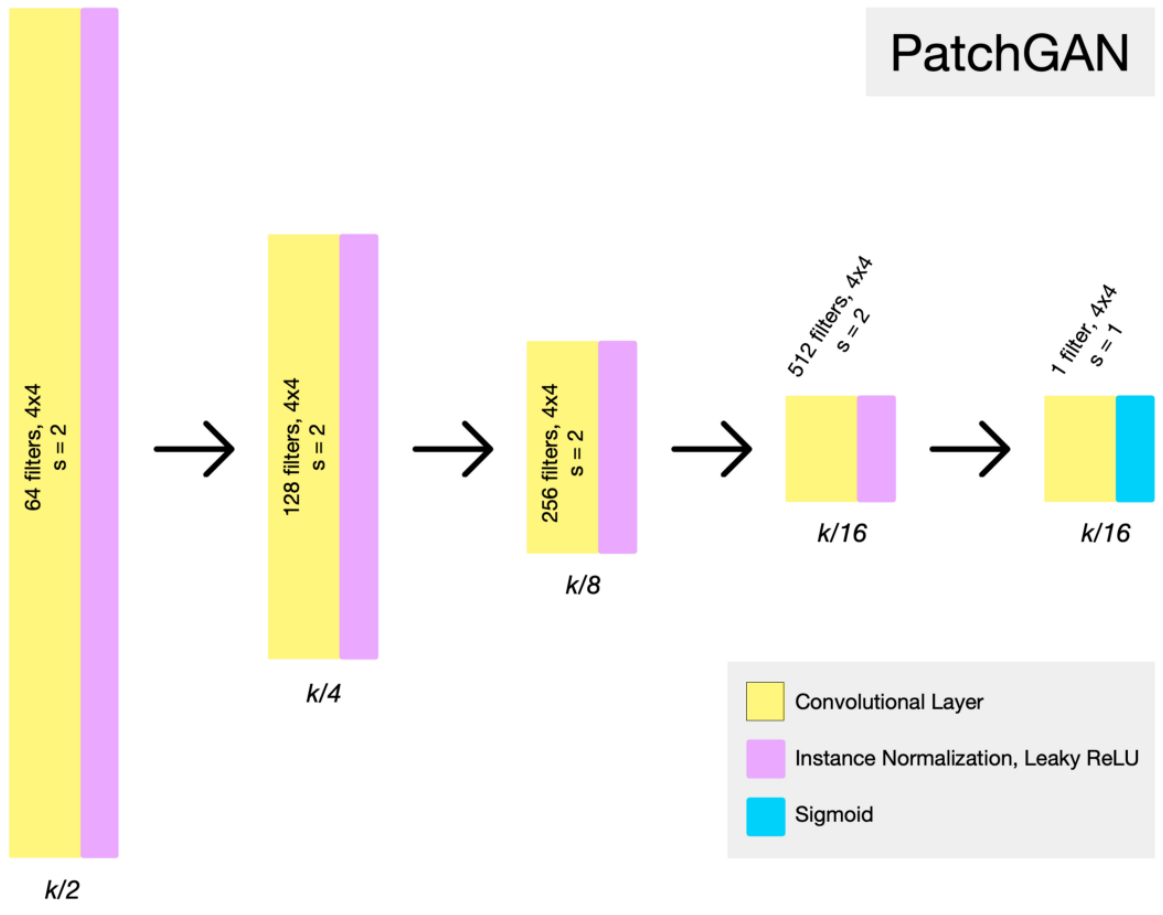}
\caption{Discriminator architecture}
\label{dis}
\end{figure}

For the discriminator networks we use 70 × 70
PatchGANs \cite{li2016precomputed,ledig2017photo}, which aim to classify whether
70 × 70 overlapping image patches are real or fake. PatchGAN is a fully convolutional network, that takes in an image, and produces a matrix of probabilities, each referring to the probability of the corresponding “patch” of the image being “real” compare to generated image. The representation size that each layer outputs is in terms of the input image size, k in Fig \ref{dis}. On each layer is listed the number of filters, the size of those filters, and the stride.  The PatchGAN halves the representation size and doubles the number of channels until the desired output size is reached.

\section{Experimental setup}
\label{exp}

We have divided the whole setup into three parts: data
collection, image processing, and model learning. 

\subsection{Dataset collection}
\begin{figure}[ht]
\centering
\includegraphics[width=\linewidth]{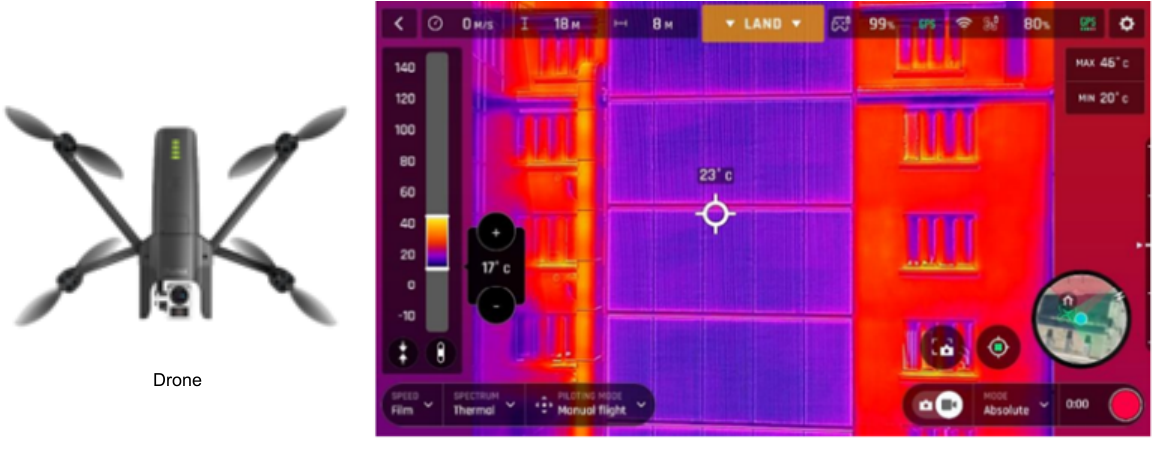}
\caption{ANAFI drone: thermal parameter setting}
\label{drone}
\end{figure}

We have collected RGB and thermal data using Parrot ANAFI Thermal \cite{anafi} drone shown in figure \ref{drone}. It  uses a infrared sensor and equipped with a 3-axis gimbal-mounted 12-megapixel camera. The camera has a 1/2.300 CMOS sensor which, in combination with a lens with a 20 mm (35 mm format equivalent) focal length.The UAV makes use of the GPS/GLONASS positioning system in combination with a barometer and Inertial Measurement Unit (IMU). The vision system requires a surface with a clear pattern and adequate lighting between 0.3 and 3 m distance from the UAV. Under normal conditions, the intelligent flight battery provides approximately 23 min of flight time.

Before using drone, camera has been calibrated to read correct measurement while flying to outside. Then temperature is adjusted while taking image in outside. It can measure heat between 14 and 752 degrees Fahrenheit. RGB and thermal image is collected different condition of the time. we collected data during midday, evening and night from two different places at UMBC \cite{umbc}. The size of the RGB image is 1920 x 1080 px, and the size of the thermal image is 1440 x 1080 px. Dataset consist of 500 images including RGB and Thermal set.

\begin{figure*}[ht]
\centering
\includegraphics[width=\linewidth]{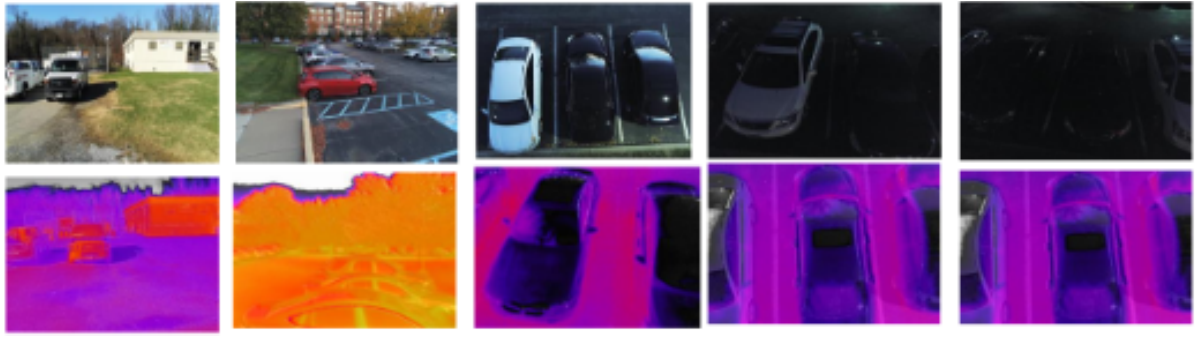}
\caption{Data sample (RGB and Thermal image ) taking during noon, evening and night time. Upper to lower,RGB image and corresponding Thermal image.}
\label{sample}
\end{figure*}

RGB related Thermal image samples are shown in Figure \ref{sample}. Temperature variation in the images are visible and different color like red or orange indicates the higher temperature variation and pink color indicates the lower temperature variation in the object.Then temperature is fixed 5 degree Celsius to 21 degree Celsius.

\subsection{Data preprocessing}

One of the major challenges in this task is the shape mismatch of the thermal image with respect to the RGB. RGB camera capture at 30 fps with shape 1920 x 1080. whereas Thermal camera capture at 9 fps with 1440 x 1080. Also, RGB image captures some in formation that is not available in thermal and vice verse. The core principle of our approach is to borrow information from the thermal image and provide this to RGB image. To tackle this problem, we have performed transformation in the both images, so that we can get the common pixels in both images.we use instance normalization \cite{ulyanov2016instance} that prevents instance-specific mean and covariance shift simplifying the learning process. Intuitively, the normalization process allows to remove instance-specific contrast information from the content image in a task like image stylization, which simplifies generation.

Resizing, Orienting, Color corrections, Augmentation \cite{shorten2019survey} such as flipping, random rotation these are  the steps taken to format images before they are used by model training. It increases  samples and model performance. Our model uses (256 x 256) size of image and we create masks and apply Portrait Mode performs on images half this size before its output is re-scaled back to full size. We applied image augmentation \cite{yang2022image} technique that create new images based on existing images in our dataset to improve the dataset as data samples are small. We applied random flips transformations that applied on all train and test samples. This transformations randomly mirror an image about its x or y axis forces the generator and discriminator to train with various image angle position from left to right or up to down. Also we used Random rotations augmentation method so that image can be used in non-fixed position.

\subsection{Training Details}

We apply least-squares loss as an objective loss which is more stable during training \cite{mao2017least}. Discriminator acts as a binary classification which differentiate the generated image and real image. The loss of discriminator and generator tracked using means square loss. I keep an image buffer that stores the 50 previously created images and set lambda = 10 that controls the relative importance of the generator and discriminator loss. Lambda identity controls the generated image with respect to real image.We update the discriminators using a history of generated images rather than the ones produced by the latest generators taking strategy from Shrivastava et al. \cite{shrivastava2017learning}.

GPU and CPU loading parameters are controlled by number of workers. The more number of workers means data load immediately and lower waiting time in GPU. We used Adam optimizer \cite{kingma2014adam} which is an adaptive learning rate algorithm requires less memory.The batch size was set to 1, which is why we refer to instance normalization, rather than batch normalization.we trained model from scratch with a learning rate of 0.001. The number of training image is 400 compare to test set which is 70.  Model
Hyper parameters that are used are given below.
\begin{itemize}
    \item Input size = 256*256px
    \item Training Epochs = 100
    \item Input batch size = 1
    \item Classification loss = MSE
    \item learning rate = 0.001
    \item Num workers = 4
    \item Lambda identity = 0.0
    \item Lambda cycle = 10
\end{itemize}

\section{RESULTS AND DISCUSSION}
\label{res}

\begin{figure}[ht]
\centering
\includegraphics[width=\linewidth]{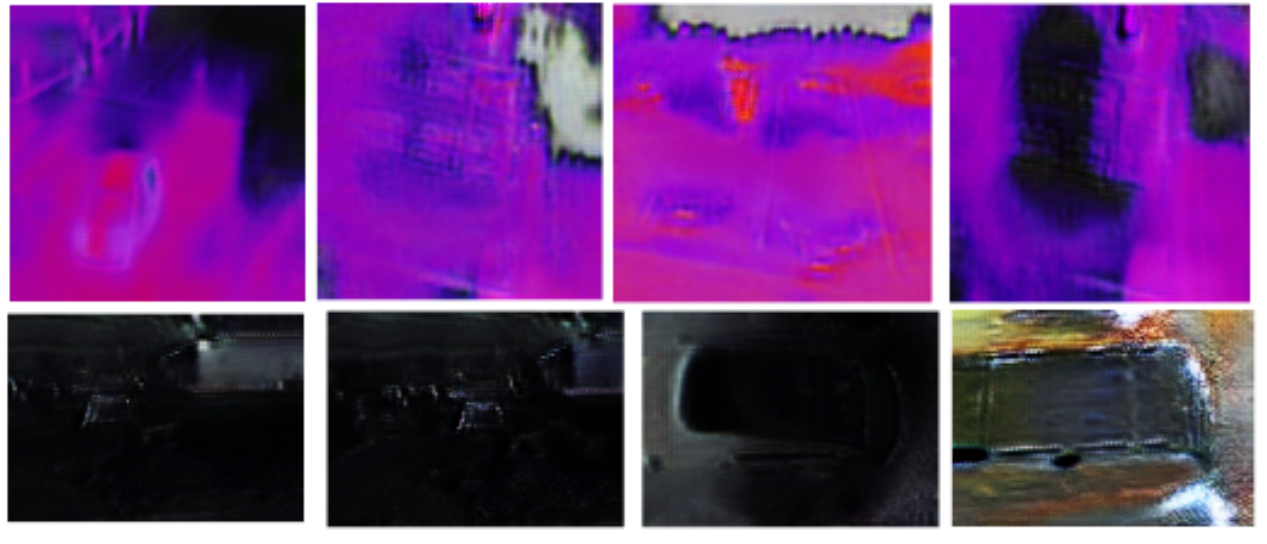}
\caption{Results of RGB to Thermal image translation for varying lighting scenarios.}
\label{genrated}
\end{figure}

Qualitative results in Figure \ref{genrated} shows generated images with respect to real image. The translated images are a little bit similar to the original RGB which indicates shortage of enough data.
The quantitative performance of model can be understood from Table \ref{restable}. MSE is used to calculate the combined square error of generated images.By reducing the loss, Generator is trying to fool the discriminator. The lower MSE loss of generator means produced image is real that are not able to differentiate by discriminator. It also refers generated image is similar to real thermal image. Discriminator MSE is increasing with training epoch which means it cannot separate real image and generated image produced by generator.

\begin{table}[ht]
\centering
\caption{Generator and Discriminator lossess}
\label{restable}
\begin{tabular}{|c|c|}
\hline

Generator\_loss & Discriminator\_loss \\ \hline
0.415           & 0.547               \\ \hline
0.358           & 0.605               \\ \hline
0.339           & 0.631               \\ \hline
0.349           & 0.646               \\ \hline
0.334           & 0.662               \\ \hline
0.329           & 0.671               \\ \hline
0.328           & 0.669               \\ \hline
0.288           & 0.693               \\ \hline
0.246           & 0.742               \\ \hline
\end{tabular}
\end{table}
Overall from Table \ref{restable}  we can observe that by reducing the loss, Generator is trying to fool the discriminator.

\section{Conclusion}
\label{con}

In this work, translations of RGB to thermal image using GAN have been investigated where data is collected by drone. I hypothesize translated thermal images will be similar to actual thermal image which can be used to do visual task in low light conditions without using any special expensive equipment for imaging. Our generated thermal image reflects the shortage of data sample but loss indicates generated image is similar to real image. Overall, the results show that it is possible to use GAN to translate from RGB to Thermal training data. This opens up for quicker and cheaper generation of thermal data, which is relevant in security and surveillance applications.

\section{Limitation and Future research}
\label{limit}

\begin{figure}[ht]
\centering
\includegraphics[width=\linewidth]{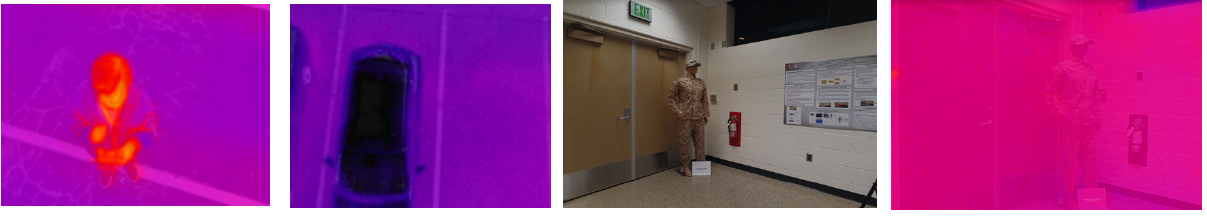}
\caption{Temperature variation in human, car and Mannequin}
\label{limitation}
\end{figure}

The study concludes that training of the model using car images alone moved the generated thermal images due to shortage of enough data. For this observation, we can the future work related to work is to expand the collection of data with different objects. Because Figure \ref{limitation} shows object variation at night. We observe that car thermal variation is lower than Human being (left one). Mannequin's RGB and thermal image  from the Figure \ref{limitation} also don't show much thermal variation. The car image has taken from out door at night whereas Mannequin's image has taken from indoor environment. But human shows the clear thermal variation even in outside environment. Another limitation of this project is model computation time. Model takes almost 2hr to complete 100 epoch. It remains for future work to try different training settings to optimise the system. Finally, the most important aim related to project is analysis loss function-cycle consistency loss alone, adversarial loss alone and compare with other image translation model.

\section{Acknowledgement}
We would like to thank Professor Dr.Nirmalya Roy for his great support and advising throughout this research.

\bibliographystyle{unsrt}
\bibliography{bib1}
\end{document}